\documentclass[]{interact}
\usepackage{graphicx}  
\usepackage{booktabs}  
\usepackage{epstopdf}
\usepackage{subfigure}
 \usepackage{graphicx}
 
\usepackage[section]{placeins}

\usepackage{colortbl}
\usepackage[table,xcdraw]{xcolor}
\usepackage{graphicx}
\usepackage{tabularx}
\usepackage{float} 
\usepackage{lmodern}   
\usepackage[T1]{fontenc}

\usepackage{array}
\usepackage[numbers]{natbib}
\usepackage{pdfpages} 
\usepackage{enumitem}
\setlist[itemize]{leftmargin=*, noitemsep, topsep=0pt}
\usepackage{multirow}
\usepackage{graphicx}
\usepackage{ltablex}   
\usepackage{multirow}  
\usepackage{array}     
\usepackage{booktabs}  
\keepXColumns          
\newcolumntype{Y}{>{\raggedright\arraybackslash}X} 
\usepackage{tabularx} 

\usepackage[normalem]{ulem}
\useunder{\uline}{\ul}{}

\usepackage{color}
\usepackage{tabularray}

\theoremstyle{plain}

\theoremstyle{definition}

\theoremstyle{remark}

\usepackage[hyphens]{url}
\usepackage[hidelinks]{hyperref}
\usepackage{url}

\usepackage[most]{tcolorbox}
\usepackage{xcolor}
\usepackage{ragged2e}

\usepackage{caption}
\usepackage{hyperref}

\definecolor{headblue}{RGB}{28,102,170}
\definecolor{headviolet}{RGB}{120,82,170}
\definecolor{headteal}{RGB}{14,120,100}
\definecolor{framegray}{RGB}{70,70,70}

\newtcolorbox{SectionBox}[2][]{enhanced,breakable,
  colback=white,
  colframe=framegray!65,
  colbacktitle=framegray!12,
  coltitle=black,
  fonttitle=\bfseries,
  boxsep=2.5pt,left=7pt,right=7pt,top=7pt,bottom=7pt,
  title={#2},#1}

\newtcolorbox{PromptCard}[2][]{enhanced,breakable,
  colback=white,
  colframe=#2!70!black,
  colbacktitle=#2!10,
  coltitle=black,
  fonttitle=\bfseries,
  boxsep=2.5pt,left=7pt,right=7pt,top=7pt,bottom=7pt,
  title={#1}}

\begin{document}

\title{Extracting Disaster Impacts and Impact Related Locations in Social Media Posts Using Large Language Models}

\author{
\name{Sameeah Noreen Hameed\textsuperscript{a}\thanks{CONTACT Sameeah Noreen Hameed Email: snoreen1@massey.ac.nz}, Surangika Ranathunga\textsuperscript{a},  Raj Prasanna\textsuperscript{a}, Kristin Stock\textsuperscript{a}, and Christopher B. Jones\textsuperscript{b}}
\affil{\textsuperscript{a}School of Mathematical and Computational Sciences, Massey University, Auckland, New Zealand; \textsuperscript{b}School of Computer Science and Informatics, Cardiff University, Cardiff, UK}
}

\maketitle

\begin{abstract}
Large-scale disasters can often result in catastrophic consequences on people and infrastructure. Situation awareness about such disaster impacts generated by authoritative data from in-situ sensors, remote sensing imagery, and/or geographic data is often limited due to atmospheric opacity, satellite revisits, and time limitations. This often results in geo-temporal information gaps. In contrast, impact-related social media posts can act as “geo-sensors” during a disaster, where people describe specific impacts and locations. However, not all locations mentioned in disaster-related social media posts relate to an impact. Only the impacted locations are critical for directing resources effectively. e.g., \textit{``The death toll from a fire which ripped through the Greek coastal town of \#Mati stood at 80, with dozens of people unaccounted for as forensic experts tried to identify victims who were burned alive \#Greecefires \#AthensFires \#Athens \#Greece.''} contains impacted location \textit{``Mati''} and non-impacted locations \textit{``Greece''} and \textit{``Athens''}. This research uses Large Language Models (LLMs) to identify all locations, impacts and impacted locations mentioned in disaster-related social media posts. In the process, LLMs are fine-tuned to identify only impacts and impacted locations (as distinct from other, non-impacted locations), including locations mentioned in informal expressions, abbreviations, and short forms. Our fine-tuned model demonstrates efficacy, achieving an F1-score of 0.69 for impact and 0.74 for impacted location extraction, substantially outperforming the pre-trained baseline. These robust results confirm the potential of fine-tuned language models to offer a scalable solution for timely decision-making in resource allocation, situational awareness, and post-disaster recovery planning for responders. 
\end{abstract}


\section{Introduction}

Disasters, triggered by natural hazards, are a major challenge that can have a devastating impact on the lives of people and infrastructure. Effective disaster management and response rely heavily on the availability of accurate information, delivered in a timely manner and appropriate format \cite{dong_social_2021}. Authoritative data, derived from sources such as in-situ sensors, remote sensing imagery, and geographic information systems (GIS) are vital for performing accurate damage assessments and coordinating relief operations \cite{ahadzadeh_earthquake_2021}. However, the acquisition of this sensor-based information is often constrained by real-world limitations, such as atmospheric opacity, which can obscure satellite views, or the inherent revisit time limitations of satellites \cite{zakeri2024synthesizing, sigopi2024advancements}. These constraints frequently lead to critical geo-temporal gaps, where timely data is unavailable \cite{al2013major, samadzadegan2025critical}. To address this information deficit and improve situational awareness, studies utilise alternative data sources such as social media text \cite{vongkusolkit_situational_2021}. Social media with its vast user base and real-time communication capabilities, has emerged as a powerful tool for gathering and disseminating critical information during disasters \cite{scalia_cime_2022}. Platforms such as Twitter, Facebook, and Instagram allow users to share first-hand observations and updates about ongoing events, effectively transforming social media into a network of Volunteered Geographic Information (VGI) \cite{sui2011convergence}. These platforms serve as a ``geo-sensor'' network, providing valuable real-time data that can complement traditional remote sensing methods and enhance situational awareness for emergency services. Social media platforms are more responsive with a more continuous flow of information during disaster times than official data, which comes in intervals that are often too long \cite{shan2021real}. 

\begin{figure}[ht]
    \centering
        \captionsetup{justification=centering, singlelinecheck=false}
    \includegraphics[width=1
    \textwidth]{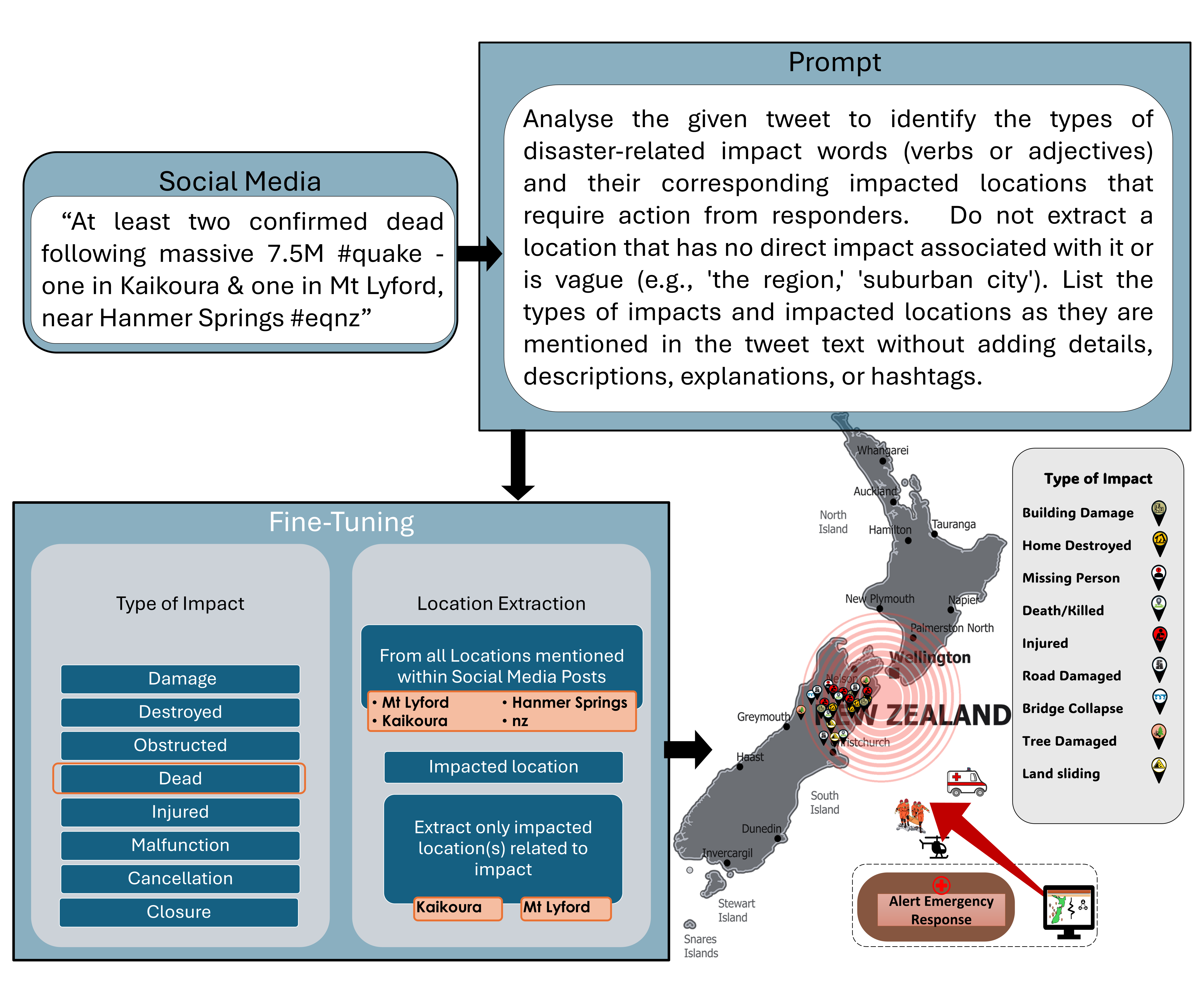}
    \caption{LLM-based real-time disaster impact mapping from social media data. The process filters non-impact locations to help emergency managers target and prioritise response efforts.}
    \label{fig:impact_loc_llms_mapping}
\end{figure}
However, the sheer volume of information generated on social media during disasters presents its own set of challenges. One of the key challenges is the identification and extraction of relevant geographic locations, or toponyms (geographically locatable place names), from the vast stream of posts \cite{suwaileh2022disaster}. These locations are critical for mapping the spatial extent of a disaster and understanding its impact. Situational awareness in disaster scenarios hinges on the ability to accurately identify the locations most affected by the event and the specific nature of the impact at each location \cite{seppanen_shared_2015}. Social media platforms generate a wealth of geospatial information that can be leveraged to geolocate disaster impacts with useful impact-related descriptions, and inform response efforts \cite{zhang_semiautomated_2020}.

Despite the potential of social media as a source of real-time geographic information, existing research in the field of location recognition or place name identification has predominantly focused on identifying a set of locations mentioned in disaster-related posts \cite{kumar2022deep,suwaileh2022disaster, suwaileh_idrisi-d_2023a,sun2025galloc}. In this paper, this set of locations mentioned in disaster-related posts is referred to as \textbf{all locations} and Twitter/X posts as \textbf{tweets}. The all locations set can also include locations that are not directly related to the disaster's impact, but rather refer to governments, organisations or other facts or events.

For instance, multiple locations are mentioned in the following social media post, \textit{``Jatlaan canal embankment is damaged. It is one of the main canals originating from Mangla dam which irrigates Punjab. \#earthquake AJK Mirpur''}. These locations include \textit{Jatlaan canal}, \textit{Mangla dam}, \textit{Punjab}, \textit{AJK}, and \textit{Mirpur}. However, only the \textit{Jatlaan canal} and \textit{AJK Mirpur} are directly impacted, as indicated by the reference to the impact \textit{damaged} and \textit{earthquake}. The presence of multiple locations can lead to confusion and inefficiencies in disaster response, as responders may struggle to identify which location requires immediate attention.

For emergency responders, different impact types require different response strategies \cite{khan2023systematic}, such as sending rescue teams with medical aid for the injured versus deploying maintenance teams for infrastructure recovery. Existing disaster classification approaches, such as that developed by \citet{alam_humaid_2021}, categorise posts according to their content. However, they do not identify the impact and the severity, to help emergency responders towards focused relief operations. For example, \citet{alam_humaid_2021} classify the following post under the infrastructure-and-utility-damage category \textit{``211 houses have been completely destroyed and a further 234 are badly damaged by the devastating \#wildfires that ripped through a coastal town east of \#Athens, killing at least 80 people. \#AthensFires \#Greecefires \#Greece"}. However, in this example, three distinct impacts: the main impact (wildfires), infrastructure damage and casualties all appear. So, there is a critical need for techniques that can accurately distinguish fine-grained disaster impacts in social media posts. This leads us to our research question: \textit{How can we identify impacts and distinguish locations directly impacted by disasters in social media postings from all locations mentioned, enabling rapid response efforts?}

 This research attempts to answer the above question by developing corpora and machine learning (ML)-based methods to accurately identify the types of impact and the directly impacted location(s) mentioned in the social media posts. The study aims to extract disaster impacts and impacted locations that can contribute to situational awareness and effective disaster response efforts during a disaster.

The contribution of this paper is two-fold.
\begin{itemize}
    \item Firstly, we develop a benchmark corpus that identifies impacted locations and types of impact in disaster-related social media text.
\end{itemize}

\begin{itemize}
    \item Secondly, building on the finding that Large Language Models (LLMs) succeeded for all locations recognition, we develop a novel methodology to identify directly impacted locations and types of impact mentioned in disaster-related social media text.
\end{itemize}

The remainder of this paper is organised into the following sections. Section \ref{Sec2:related-work} discusses related work and previous approaches. Section \ref{Section-3:dataset} highlights the details of the dataset used in this research. It is followed by the detailed methodology in Section \ref{sec4: methodology} and the experimental setup in Section \ref{Sec-5:ExperimentalSetup}. The results and their analysis are presented in Section \ref{sec6-results}. Section \ref{sec7-discussion} discusses the key findings and limitations. Finally, Section \ref{sec8-conclusion} concludes the paper and outlines directions for future work. 

\section{Related Work} \label{Sec2:related-work}

\subsection{Impact Information in Disaster Posts}

Disasters such as earthquakes, hurricanes, floods, and wildfires are catastrophic events that disrupt communities and ecosystems, resulting in loss of life, destruction of property, and long-term economic and social impacts \cite{chaudhary2021natural}. Social consequences include population displacement \cite{lindell_assessing_2003}, shortages of essential resources \cite{ragini2018big}, and psychological effects such as anxiety and trauma \cite{lee2020water, charnley2021traits}. 

While social and economic research examines human and financial impacts \cite{arcaya2020social, hallegatte2015indirect}, the most visible impacts are related to physical damage, and are frequently reported via real-time social media posts \cite{alam_humaid_2021}.  Physical impacts such as casualties, infrastructure damage, and service outages demand urgent attention from emergency responders. These impacts are often captured promptly on social media through user-generated posts containing text, images, or videos of affected areas \cite{zhang2020semiautomated, huang2015geographic, xie2018using}. Unlike economic losses that unfold over time \cite{berariu2015understanding}, physical damage information shared online is highly actionable during the early response phase. 

\citet{xiao_understanding_2015} and \citet{shan2021real} demonstrated that impact information in social media posts, such as casualties, infrastructure damage, and economic losses, enhances situational awareness and rapid response planning. Despite this, limited research has explored the extraction of fine-grained types of impact information along with locations.

\subsection{Approaches to Location Extraction}
\label{SubSec:Approcaches_location_extraction}
Early research on location recognition in social media used user profile or geo-tag information from Twitter/X to infer the tweet’s origin \citep{hecht_tweets_2011, ikawa_location-based_2013}. However, a user's profile location does not necessarily reflect their real-time physical location when a tweet is posted, and profile-based locations can be inaccurate or irrelevant \citep{stock_mining_2018}. This led to approaches focusing on identifying locations mentioned in the tweet text itself. \citet{imran2015processing} explored how points of interest can be identified based on the locations mentioned in the text.

Various approaches and techniques have been used to improve the precision of location recognition strategies and algorithms. Named Entity Recognition (NER) is one of the widely used approaches for location recognition from social media posts \cite{zhang_semiautomated_2020}. Some of the commonly used NER tools for location recognition include the use of Stanford NER \cite{finkel_incorporating_2005} and SpaCy-NER\footnote{SpaCy-NER available at: \url{https://spacy.io/universe/project/video-spacys-ner-model}}. Both of these tools have been used by researchers for geoparsing, which involves location recognition and resolution from unstructured text into geographic coordinates, in disaster-related tweets \cite{dutt_savitr_2018, lingad_location_2013}. 

Despite the availability of these NER tools, certain challenges impact their performance in location extraction from social media posts. Some of the major challenges include the use of abbreviations, incorrect spellings, and slang in social media posts \cite{liu2023improving}. Another challenge with NER tools is that they cannot reliably distinguish between locations and similar named entities (e.g., \textit{Washington State} vs. \textit{Washington, D.C.} vs. \textit{George Washington} (person) \cite{seow2025review}. Moreover, most of these NER tools are trained to recognise multiple entities (e.g. people and organisations), rather than focusing specifically on location entities, which can reduce their reliability for this task.

To overcome the limitations of publicly available, generic NER tools, researchers have devised mechanisms that explicitly focus on location extraction. These approaches include traditional Machine learning (ML) and Deep Learning (DL) techniques. Early examples of ML approaches, such as Support Vector Machines (SVMs), Naïve Bayes (NB), and Random Forests (RF) have been applied to location extraction from social media, using lexical, grammatical, and geographical features with the Beginning, Inside, Last, Unit-length (BILOU) schema and rule-based methods to enhance performance \citep{li_fine-grained_2014, hoang_location_2018}. DL techniques greatly advanced location extraction by capturing complex patterns in data through a range of Neural Network based methods. These include a Convolutional Neural Network (CNN) \cite{kumar_location_2019}, the Bidirectional Long Short-Term Memory (Bi-LSTM) model, which enhanced the recognition  of rare or unknown locations from Twitter/X \cite{chen_location_2021} and a fusion of CNN and LSTM model for geoparsing city-level geolocation \cite{mahajan_predicting_2021}. 

Transformer models, particularly BERT\cite{devlin_bert_2019}, revolutionised location recognition through contextual embeddings. \citet{suwaileh2022disaster} reported that BERT-based models achieved superior performance in location recognition and disambiguation(differentiating similar locations such as Paris, France vs Paris, Texas) tasks, surpassing all baseline models such as BiLSTM-CRF and SpaCy-NER. \citet{fan2020hybrid} proposed a hybrid approach for disaster situational awareness that combined NER for location detection, a location fusion method for refining coordinates, and a fine-tuned BERT model for classifying posts using Hurricane Harvey data. Transformer-based models like BERT set the benchmark with strong contextual understanding; however, their state-of-the-art (SOTA) performance in general entity recognition diminishes in the noisy, time-critical context of disaster response, where they often fail to accurately extract fine-grained, multi-level location information \cite{suwaileh2022disaster,hu_geo-knowledge-guided_2023}.

Recent advances integrate LLMs for fine-grained spatial reasoning. \citet{hu_geo-knowledge-guided_2023} introduced geo-knowledge-guided prompting for detailed location descriptions in disaster texts. \citet{yin2025llm} use LLM-based techniques to process large-scale textual and visual data for implicit geospatial information, improving the interpretation of location details. \citet{sun2025galloc} developed the annotation tool GALLOC, for labelling and handling detailed, multi-level location descriptions in disaster-related texts. 
LLMs have moved beyond traditional sequence labelling and shifted towards generative and few-shot learning capabilities, which enable them to perform NER as a text-to-text generation problem \cite{ashok_promptner_2023, wang_gpt-ner_2023}. Researchers used various prompting techniques, such as Chain-of-Thought \cite{wei2022chain} and Retrieval-Augmented Generation (RAG), to enhance performance and deal with hallucinations, where the model invents non-existent locations or geospatial facts \cite{dhuliawala2023chain,huang2025survey,yin2025llm,xu2024large}. Early prompting-based studies found ChatGPT\footnote{OpenAI. ChatGPT. Available at: \url{https://chatgpt.com/}} could perform well in zero to few-shot settings; however, it underperformed relative to fine-tuned transformer baselines on CoNLL-2003 and OntoNotes \cite{wang2023gpt}. \citet{cheng_novel_2024} standardised prompts with four elements in few-shot settings for the NER task, along with self-check and correction implementation at the end in output generated by LLMs. Their experiments demonstrated that LLMs still encounter difficulties in technical fields involving specialised terminology in entity recognition. \citet{zaratiana_gliner_2024} showed that the GLiNER model outperformed LLMs in zero-shot and low-resource NER tasks, due to format sensitivity and lack of structured span control in LLMs. Studies have also shown that while general-purpose LLMs possess a surprising amount of geospatial knowledge \cite{kopanov_comparative_2024}, fine-tuning with domain-specific data and incorporating geo-knowledge graphs can significantly improve the accuracy in the location extraction task from disaster-related social media messages \cite{sun2025galloc, hu_geo-knowledge-guided_2023}. demonstrates how LLMs can integrate textual and visual cues to infer location context explicit location mentions. However, LLMs consistently face challenges due to hallucination \cite{ji2023survey,kalai2025language}, sensitivity to prompt formation, and the presence of long contexts \cite{liu2023pre}.

Existing studies have made significant progress using traditional NLP models, transformer-based architectures and more recently, LLMs for identifying location from social media posts. However, these studies focused on identifying all locations mentioned within a post, without considering whether these locations are directly affected by the disaster impact being described. This highlights a critical gap that this research addresses by focusing on the extraction of impacted locations and the impacts to enable actionable situational awareness for disaster response.

\subsection{Classification of Disaster-related Social Media Posts}
\label{SubSec:Classification_of_disaster}
The task of extracting disaster-related information from social media text has often been approached as a classification task. Research in this area has focused on two dimensions: (1) whether a post is relevant to the disaster or not, and (2) categorising posts based on the content.
\subsubsection{Taxonomy of Disaster-related Categories}
\label{sec2.1.1-taxonomy-disaster-categories}

Many studies have focused on binary classification of disaster-related posts which distinguishes between relevant/irrelevant or informative/uninformative posts.
\citet{imran_may_2013} presented one of the early efforts towards the classification of disaster-related posts. Their framework has two stages: classifying tweets as \textit{informative} or \textit{non-informative}, followed by categorisation into specific types such as \textit{caution/advice}. In addition to work identifying whether a post is relevant to a disaster, or informative, there have been a number of studies that have presented detailed taxonomies  \cite{ashktorab_tweedr_2014, parilla-ferrer_automatic_2014}. \citet{olteanu2014crisislex} begin with a binary classification, which was later extended in \cite{olteanu2015expect} to classify posts into six humanitarian categories. \citet{caragea_classifying_2011} classified SMS messages from the Haiti earthquake into 10 categories. \citet{imran_twitter_2016} and \citet{alam_humaid_2021} further developed the disaster-related posts classification on 19 different disasters across the globe. \citet{alam_humaid_2021} refined the categories defined by them in \cite{imran_twitter_2016} to include \textit{requests/urgent needs} and a \textit{don’t know/can’t judge} category for non-English or ambiguous posts.
Building on these foundational efforts, recent work has shifted focus from relevance to detecting actionable tweets. For example, \citet{kruspe2021actionable} highlighted the necessity of distinguishing actionable tweets from relevant tweets. \citet{gao2024knowledge} propose a knowledge injected prompt learning method rather than using standard classifier pipelines, into actionable categories rather than just \textit{relevant/irrelevant}. \citet{lamsal2025actionable} also highlights tweets classified as \textit{requests or offers} often convey support, generic behavioural guidance, or outdated information, and therefore subdivide this class into four subcategories: \textit{Request, Offer, Irrelevant, Request and Offer}. Table \ref{tab:classification_disaster_posts} summarises the classification efforts by various researchers.

\begin{table}[htbp]
 \centering
    \captionsetup{justification=centering, singlelinecheck=false}
 \caption{Classification of disaster-related social media posts}\label{tab:classification_disaster_posts}
 \resizebox{1.1\columnwidth}{!}{%
\begin{tabular}{|l|l|l|l|}
\hline
\textbf{Paper}                                       & \textbf{Classification}                                                                                                                                                                                                                                                                                                     & \textbf{Methodology}                                                                                                                                                           & \textbf{Techniques}                                                       \\ \hline
\citet{imran_may_2013}              & \begin{tabular}[c]{@{}l@{}}Informative vs Non-informative,\\ Caution and advice, Information\\ source, Donations, Causalities\\ \& damage, Unknown\end{tabular}                                                                                                                                                             & \begin{tabular}[c]{@{}l@{}}Used traditional machine learning \\ algorithms (NB, SVM) for \\ classification.\end{tabular}                                              & ML Techniques                                                             \\ \hline
\citet{ashktorab_tweedr_2014} & Related vs. Not related                                                                                                                                                                                                                                                                                                     & \begin{tabular}[c]{@{}l@{}}Built a pipeline to identify disaster-related\\ tweet relevancy focusing on damage and\\ casualties.\end{tabular}                                   & \begin{tabular}[c]{@{}l@{}}Rule-Based \\ Techniques\end{tabular}          \\ \hline
\citet{olteanu2014crisislex}          & Informative vs Uninformative                                                                                                                                                                                                                                                                                                & \begin{tabular}[c]{@{}l@{}}Manual annotation of crisis tweets followed \\ by lexicon construction using frequent term\\ extraction and distributional similarity.\end{tabular} & \begin{tabular}[c]{@{}l@{}}Lexicon-Based, \\ Semi-Supervised\end{tabular} \\ \hline
\citet{parilla-ferrer_automatic_2014} & Informative vs Non-informative                                                                                                                                                                                                                                                                                              & \begin{tabular}[c]{@{}l@{}}Developed a framework to classify tweets \\ based on content relevancy.\end{tabular}                                                                & ML Techniques                                                             \\ \hline
\citet{nguyen_robust_2017}           & Relevant vs. Irrelevant                                                                                                                                                                                                                                                                                                     & \begin{tabular}[c]{@{}l@{}}Employed Convolutional Neural Networks\\ (CNNs) for tweet classification, achieving\\ higher accuracy.\end{tabular}                                 & \begin{tabular}[c]{@{}l@{}}Deep Learning \\ Techniques\end{tabular}       \\ \hline
\citet{rathod_disaster_2022}        & Relevant vs. Irrelevant                                                                                                                                                                                                                                                                                                     & \begin{tabular}[c]{@{}l@{}}Utilized Word2Vec-based language \\ embeddings with machine learning models\\ for classification.\end{tabular}                                      & ML Techniques                                                             \\ \hline
\citet{olteanu2015expect}             & \begin{tabular}[c]{@{}l@{}}Affected individuals, \\ Infrastructure and utilities,\\ Donations and volunteering,\\ Caution and advice, Sympathy\\ and emotional support, Other \\ useful information\end{tabular}                                                                                                            & \begin{tabular}[c]{@{}l@{}}Manually annotated tweets from 26 crisis\\ events into six humanitarian categories \\ using crowdsourcing and expert review.\end{tabular}           & Manual                                                                    \\ \hline
\citet{caragea_classifying_2011}      & \begin{tabular}[c]{@{}l@{}}Medical emergency, People \\ trapped, Shelter needed,\\ Food distribution, Food\\  shortage, Water shortage, \\ Water sanitation, Collapsed\\  structure, Hospital services, \\ Person news\end{tabular}                                                                                         & \begin{tabular}[c]{@{}l@{}}Classified SMS messages from the Haiti \\ earthquake into specified categories based \\ on content.\end{tabular}                                    & ML Techniques                                                             \\ \hline
\citet{imran_twitter_2016}           & \begin{tabular}[c]{@{}l@{}}Injured/dead people, Sympathy/\\ support, Displaced people/\\ evacuations, Infrastructure\\ damage, Donation needs/offers, \\ Missing/trapped/found people,\\ Caution/advice, Not related,\\ Other useful\end{tabular}                                                                           & \begin{tabular}[c]{@{}l@{}}Developed a classification strategy based\\ on manually annotated tweets from 19 \\ different disasters.\end{tabular}                               & ML Techniques                                                             \\ \hline
\citet{alam_humaid_2021}              & \begin{tabular}[c]{@{}l@{}}Injured or dead people, Missing/\\ trapped/found people, Requests/\\ urgent needs, Displaced people/\\ evacuations, Rescue/donation/\\ volunteering, Caution/advice, \\ Sympathy/support, Infrastructure\\ damage, Other useful information, \\ Not related, Don’t know/can’t judge\end{tabular} & \begin{tabular}[c]{@{}l@{}}Improved previous classifications by \\ splitting and refining categories based\\ on tweet content.\end{tabular}                                    & ML Techniques                                                             \\ \hline
\citet{suwaileh2023idrisi-re-2023b}     & \begin{tabular}[c]{@{}l@{}}Same categories as \\ \citet{alam_humaid_2021} \end{tabular}                                                                                                                                                                                                                                            & \begin{tabular}[c]{@{}l@{}}Further refined classification strategies\\ for disaster-related tweets.\end{tabular}                                                               & ML Techniques                                                             \\ \hline
\citet{gao2024knowledge} & 
\begin{tabular}[c]{@{}l@{}}Advice, Threat, Service, Volunteer\end{tabular} & 
\begin{tabular}[c]{@{}l@{}}Proposed a knowledge-injected \\ prompt learning approach to enhance \\ actionable category detection.\end{tabular} & 
\begin{tabular}[c]{@{}l@{}}Transformer-based \\ Deep Learning\end{tabular} \\ \hline

\citet{lamsal2025actionable} & 
\begin{tabular}[c]{@{}l@{}}Request, Offer, Irrelevant, \\ Request and Offer categories\end{tabular} & 
\begin{tabular}[c]{@{}l@{}}Classification of tweets into \\ actionable categories.\end{tabular} & 
\begin{tabular}[c]{@{}l@{}}Transformer-based \\ Classification\end{tabular} \\ \hline

\end{tabular}
 }
 \end{table}

\subsubsection{Techniques for Classifying Disaster Posts }
Various techniques have been proposed for automatically categorising social media posts based on the taxonomies mentioned in Section \ref{sec2.1.1-taxonomy-disaster-categories}. Some early works, such as \citet{ashktorab_tweedr_2014}, used rule-based techniques to classify posts based on content relevancy to disaster keywords. ML approaches were widely used in research that treated the task as a supervised binary classification problem \cite{imran_may_2013, parilla-ferrer_automatic_2014} or as a multi-class classification problem for more fine-grained information \cite{caragea_classifying_2011}. These approaches mainly focused on the bag-of-words technique with NB and SVM methods. There have been some efforts using Word2Vec-based language embeddings and machine-learning models for both binary \cite{rathod_disaster_2022} and multi-class classification \cite{imran_twitter_2016} tasks. \citet{nguyen_robust_2017} employed deep learning-based CNNs to classify tweets as \textit{relevant} or \textit{irrelevant}, achieving higher accuracy compared to traditional methods.

Transformer-based architectures have significantly advanced disaster posts classification. Pre-trained language models such as BERT and RoBERTa have been fine-tuned for disaster-related tasks, demonstrating substantial improvements in capturing contextual meaning beyond bag-of-words or Word2Vec embeddings \cite{alam_humaid_2021,lamsal2024crisistransformers}. Models specifically adapted for crisis informatics, such as CrisisBERT \cite{liu2021crisisbert}, leverage domain adaptation techniques to handle the unique vocabulary and context of disaster-related social media.

Beyond BERT-based models, larger generative models and instruction-tuned LLMs have recently been applied for the classification of disaster tweets. Zero-shot and few-shot prompting with LLMs (e.g., GPT-3, Llama, and Mistral) have shown promising performance, reducing dependence on large annotated datasets \cite{zhu2024gl,xiao_llm-named_2024,rawte2023survey}. \citet{mcdaniel2024zero} used GPT-4, Gemini, and Claude in zero-shot classification, showing LLMs outperform traditional supervised methods for both binary and other humanitarian categories. \citet{yin2025llm} proposed CrisisSense-LLM, which uses instruction fine-tuning for multi-label classification of social media posts, enabling robust detection of multiple humanitarian information types within a single tweet. \citet{imran2025evaluating} evaluated the robustness of LLMs across 19 disaster events and linguistic contexts, demonstrating their adaptability while also highlighting challenges with nuanced classes.

Overall, the transition from rule-based approaches to ML, DL, and, most recently, LLMs illustrates a sustained effort to enhance the robustness, generalizability, and timeliness of disaster tweet classification. In particular, LLMs mark a significant shift toward flexible and data-efficient methods that can be quickly adapted and deployed in response to emerging and unforeseen crisis events. However, despite these advancements, most existing approaches focus on broad categorical classifications and do not extract fine-grained information such as the specific impact or the directly impacted locations. This gap highlights the need for models capable of capturing detailed impact-level information from social media text to support more targeted and actionable disaster response.

\subsection{Disaster-Related Social Media Datasets} \label{SubSec:disaster_data}
Social media platforms such as Twitter, Facebook, and Instagram create a stream of information as soon as an event occurs. Twitter has traditionally been considered the primary medium by researchers \cite{stock_mining_2018}, due to its real-time nature and ease of sharing short, concise messages. However, since Twitter rebranded to ``X'' in 2023, major policy changes occurred, particularly the removal of free API access. Ongoing studies continue to use ``X'' data despite API policy changes for disaster research \cite{aktuna2023investigating,patel2025eq, karimiziarani2023social}.

Several disaster related social media data sets have been developed, \citet{sit_identifying_2020} extracted about 20,000 Hurricane Irma tweets and manually annotated 10,000 of them for relevance, informativeness, and situational categories to support disaster response analysis. Similarly, \citet{scheele2021geographic} collected Tweets using keywords related to Hurricane Sandy, during the timeframe of the disaster via the Twitter API. The HumAID dataset \citep{alam_humaid_2021} is one of the largest collections of disaster-related social media data, comprising tweets from 19 different disaster events, classified into nine categories of disaster-related information. However, this dataset lacks explicit annotations for locations and impact types, limiting its direct applicability to location-specific disaster impact analysis. More recently,  \citet{suwaileh2023idrisi-re-2023b}  used the HumAID dataset for the location mention recognition task (LMR) task. They used manual annotation and synthetic annotation generation techniques to develop a corpus of 20.5k humanly annotated tweets and 57k automatically labelled tweets. They further extended it for the location disambiguation task and termed it the IDRISI-D dataset, which consists of 5,591 manually annotated tweets. The English subset (IDRISI-DE) contains tweets that have one or more locations and was annotated to identify all locations mentioned in the tweet text. Each location was then manually disambiguated. However, a key limitation of IDRISI-DE is that it does not distinguish locations that were directly impacted by the disaster, which is an essential requirement for our task. Also, it does not focus on the types of impacts mentioned in the disaster tweets. Table \ref {tab:disaster_datasets} summarises the key attributes of the above-described data sets.

\begin{table}[H]
\centering
    \captionsetup{justification=centering, singlelinecheck=false}
\caption{Key Twitter datasets on disaster management}
\label{tab:disaster_datasets}
\resizebox{\columnwidth}{!}{%
\begin{tabular}{|l|l|l|l|}
\hline
\textbf{Reference} &
  \textbf{Disaster Event(s)} &
  \textbf{Data Volume} &
  \textbf{Purpose/Task} \\ \hline
 \citet{sit_identifying_2020} &
  Hurricane Irma (2017) &
  20,000 labelled tweets &
  \begin{tabular}[c]{@{}l@{}}location recognition \\ from tweets\end{tabular} \\ \hline
\citet{scheele2021geographic} &
  Hurricane Sandy (2012) &
  \begin{tabular}[c]{@{}l@{}}1,920 tweets labelled \\ informative/uninformative\end{tabular} &
  \begin{tabular}[c]{@{}l@{}}Geo-tagged tweets based \\ on keyword search\end{tabular} \\ \hline
\citet{imran_twitter_2016} &
  19 different disasters &
  50,000 labelled Tweets &
  \begin{tabular}[c]{@{}l@{}}Extracting important \\ disaster-related insights\end{tabular} \\ \hline
\citet{alam_twitter_2018} &
  \begin{tabular}[c]{@{}l@{}}Hurricane Harvey (2017), \\ Irma (2017), Maria (2017)\end{tabular} &
  50,000 labelled Tweets &
  \begin{tabular}[c]{@{}l@{}}Humanitarian categories\\ classification\end{tabular} \\ \hline
\citet{alam_humaid_2021} &
  19 disasters (2016–2019) &
  \begin{tabular}[c]{@{}l@{}}77,000 human-annotated \\ tweets\end{tabular} &
  \begin{tabular}[c]{@{}l@{}}Human-annotated dataset for \\ humanitarian classification\end{tabular} \\ \hline
\citet{imranTBCOV2022} &
  COVID-19 pandemic &
  \begin{tabular}[c]{@{}l@{}}500 manual and 1,000\\ crowd-sourced annotations\end{tabular} &
  \begin{tabular}[c]{@{}l@{}}Sentiment and location\\ recognition \end{tabular} \\ \hline
\citet{suwaileh2023idrisi-re-2023b} &
  19 different disasters &
  \begin{tabular}[c]{@{}l@{}}5,723 manually annotated \\ tweets,  25,034 automatically \\labelled\end{tabular} &
  Location extraction \\ \hline
\citet{suwaileh_idrisi-d_2023a} &
  19 different disasters &
  \begin{tabular}[c]{@{}l@{}} 3,893 tweets (manually \\ annotated) \end{tabular} &
  Location disambiguation \\ \hline
\end{tabular}%
}
\end{table}

Further, \citet{ziaullah_monitoring_2024} generated synthetic data, due to the lack of publicly available data, tagging critical infrastructure with impacts, severity, and operational statuses of limited critical infrastructure for Broward County, Florida, USA and Christchurch, New Zealand. It was generated with the help of LLMs and also focuses only on two specific disasters. 

Overall, early datasets focused on identifying disaster-related content, including humanitarian categorisation and location extraction. Recent datasets incorporated location disambiguation and automated annotation for scalability. However, none directly distinguish impacted locations or capture detailed impact information. This gap underscores the need for a fine-grained dataset with specific impacts to affected locations to enable more accurate and actionable disaster analysis.

\section{Corpus Creation for Impacted Location Recognition} \label{Section-3:dataset}

\subsection{Data Collection}

For the task of location recognition in disaster-related social media data, we selected the publicly available IDRISI-D English corpus (IDRISI-DE) \citep{rsuwaileh2024idrisid}, since it provides all location mentions and categories proposed by \citet{alam_humaid_2021} across a diverse range of disaster types. However, not all of these categories contribute towards the situational awareness of the disaster for emergency response. For example, tweets related to \textit{rescue volunteering or donation efforts} or those providing \textit{caution and advice} do not provide direct evidence of disaster impacts on people or infrastructure. Similarly, \textit{requests or urgent needs} and \textit{displace people and evacuations} are important for humanitarian coordination, but less informative for emergency response.

To address this, we selected three critical information categories from IDRISI-DE that are highly relevant to disaster response: (1) \textit{injured or dead people}, (2)\textit{ infrastructure and utility damage}, and (3) \textit{missing or found people}, which resulted in a reduction of tweets from IDRISI-DE for the annotation. We refer to this newly publicly available dataset as the Disaster Impacted Location Corpus (DILC)\footnote{DILC available at: \url{https://github.com/masseygeoinformaticscollaboratory/DILC}}.

\subsection{Corpus Annotation} 
The development of DILC\footnote{This study was evaluated under the Massey University ethical policies and procedures, was judged to be low risk and has been peer approved with approval number 4000030237)} is based on annotating the impact and impacted location mentioned within disaster-related social media posts.

The annotations provide a refined dataset that explicitly identifies disaster impacts and impacted locations, improving the accuracy and relevance of location recognition in disasters. To achieve this, a manual analysis of each tweet was performed by two annotators, who were responsible for identifying only the impact and locations that were directly impacted by the disaster.  Additionally, an author acted as a meta-annotator to resolve conflicts and ensure consistency, enhancing the qualitative credibility of the dataset. 
We crafted annotation guidelines in Appendix \ref{appendix:annotation_guidelines} that contain examples of impacted locations versus non-impacted locations and how to identify different types of impacts. We selected BRAT\footnote{BRAT is an online environment for collaborative text annotation, available at: https://brat.nlplab.org/index.html} as the annotation tool for the manual annotation process. Each annotator held at least a master's degree and possessed basic knowledge of the task.

Inter-Annotator Agreement (IAA) calculated among annotators based on Cohen's kappa for the impact was 61\%, and for impacted locations was 81\%. While the kappa score for the impact is modest, this level of agreement is consistent with prior disaster annotation studies where interpreting impact-related information introduces subjectivity \cite{khanal2021identification}. The higher agreement for the impacted locations suggests that they are easier to detect, unlike impacts that often overlap and are expressed ambiguously in social media posts.

The resulting DILC data set labels both the impact and the impacted location of 19 large disaster events of different types (e.g., wildfires, cyclones, earthquakes, floods, etc.) spanning 11 countries, covering both native and non-native English-speaking regions, and contains 1461 tweets annotated with 3359 types of impact and 1831
impacted locations out of 2649 locations mentioned.

\section{Methods for Location Recognition } \label{sec4: methodology}
In this section, we describe the methods used for location recognition, focusing on two distinct tasks: the general all location extraction task and the proposed task of identifying impacted locations. We categorise the location recognition approaches into three main types: (1) Traditional NER Tools (2) Pre-trained Language Models, and (3) Large Language Model (LLM) approaches.

\subsection{Traditional NER Tools}
The first category involved the use of traditional NER tools for location recognition, specifically SpaCy and Flair. Both tools were applied to extract all of the locations mentioned in disaster-related social media posts. SpaCy\footnote{SpaCy, available at: https://spacy.io} is a widely used NLP toolkit for NER tasks, and we utilised its \textit{en\_core\_web\_trf} model. It is the largest English model optimised for speed and accuracy in recognising named entities, including locations. SpaCy employs a transition-based dependency parser, combined with CNNs and transformer-based embeddings, enabling efficient context-sensitive token classification. Flair \cite{akbik2019flair}, on the other hand, employs contextual string embeddings and character-level language models, offering robust performance in NER tasks. On top of these embeddings, Flair employs a BiLSTM-CRF architecture: the BiLSTM captures dependencies across the sentence in both directions, while the CRF layer ensures consistent sequence labelling. It enables cases where the same word may carry different meanings in different contexts to be distinguished. Flair \textit{v0.15.0} was selected as it is the latest model that integrates character-level language, which allows it to handle noisy text, misspellings, and rare words commonly found in disaster-related social media posts.

\subsection{Pre-trained Language Models}
The second approach leveraged pre-trained language models, including BERT \cite{devlin_bert_2019}, XLM-RoBERTa \cite{conneau2020unsupervised}, and GLiNER \cite{zaratiana-etal-2024-gliner}. Both BERT and XLM-RoBERTa use a bidirectional transformer encoder, which allows them to learn context from both the left and right sides of a token at once.  Context-aware representation is particularly effective in identifying locations within disaster-related posts, where location mentions often appear in varied syntactic structures\cite{BERTNER2018}. We employed the fine-tuned BERT-NER model \textit{dslim/bert-base-NER}\footnote{BERT-NER, available at: https://huggingface.co/dslim/bert-base-NER, last accessed: 3-2-2025} trained on location data \cite{kaur2024bert}. We also experimented with XLM-RoBERTa-\textit{large-finetuned-conll03-english}\footnote{XLM-RoBERTa, available at: \url{https://huggingface.co/FacebookAI/xlm-roberta-large-finetuned-conll03-english}}, a multilingual transformer model with cross-lingual representations, fine-tuned for English NER. It is capable of handling diverse linguistic contexts and offers strong generalisation, making it particularly effective for recognising region-specific place names in local languages (e.g., \textit{Alibag, Kaikoura}) commonly found in social media data.
GLiNER\footnote{GLiNER, available at: \url{https://github.com/urchade/GLiNER}}  is a label-aware NER model that is designed to take the target entity labels (e.g., LOC, PER, ORG) as an explicit input during the prediction process, and it leverages pre-trained language encoders like XLM-RoBERTa. We employed GLiNER \textit{2.13} to extract all location entities, as it has demonstrated superior performance over LLMs in zero-shot and low-resource NER tasks \cite{zaratiana-etal-2024-gliner}.

\subsection{ Large Language Model (LLM)-based approaches}
The third approach involved the use of LLMs, an advanced form of pre-trained language models that learn contextual representations from large-scale text data using transformer architectures. These pre-trained language models are post-trained (fine-tuned or instruction-tuned) on domain data for task adaptation. In this study, we denote models as either \textbf{pre-trained} or \textbf{fine-tuned} to clearly indicate their respective training stages and usage contexts.  We experimented with several state-of-the-art LLMs for their capacity to identify locations within disaster-related social media content, with a focus initially on their effectiveness in recognising all locations mentioned. These included the following:
\begin{itemize}
\item Mistral\footnote{Mistral, available at:\url{https://huggingface.co/docs/transformers/en/model_doc/mistral}}, a 7 billion parameter model that is used for its contextual reasoning capabilities. 
\item
Qwen-2.5\footnote{Qwen 2.5 available at: \url{https://arxiv.org/abs/2412.15115v1}} with 7 billion parameters, which has strong instruction-following ability, 
multilingual coverage, and robust reasoning. 
\item
Gemma\footnote{Gemma, available at: \url{https://deepmind.google/models/}} with 2 billion parameters, which is a lightweight yet powerful model designed for efficiency and multilingual coverage, making it suitable for noisy, user-generated text. 
\item
{Phi-3 \footnote{Phi-3, available at:\url{https://huggingface.co/docs/transformers/en/model_doc/phi}}}, a medium-sized model with 14 billion parameters optimised for reasoning and factual consistency that enables the model to handle complex linguistic contexts. 
\item
{Llama-3\footnote{Llama 3, available at:\url{https://huggingface.co/meta-llama}}}, 3 billion - 70 billion parameters, which is optimised for understanding and generating human-like text.
\end{itemize}
These LLMs were tested to identify the best-performing LLM for comparison against traditional NER tools and pre-trained language models.

\subsection{Prompt Engineering}

In our LLMs experiments, we employed several prompt engineering techniques to evaluate their effect on two tasks: all location recognition and impact and impacted location recognition. The prompt types utilised in our research include Basic, Persona, and Chain-of-thought, drawing on established strategies and reasoning-based prompting \cite{brown2020language, schick2021size, wei2022chain, liu2023pre}. For each prompt type, we experimented with zero-shot, one-shot, and 6-shot settings.

We utilised ChatGPT-4 to formulate prompts, leveraging its strong instruction-following and reasoning capabilities to optimise response quality. An iterative process was applied to small sets of tweets to refine prompt instructions. To avoid non-geographic contexts (e.g., ``\textit{Greece}'' from \textit{\#prayforGreece}), we added counter-examples in the prompt \cite{wei2022emergent}, to restrict extraction to valid, standalone locations. For all location recognition, the prompt was modified to track duplicate mentions, which we resolved by enforcing structured output that required each location to be listed with its frequency \cite{cheng_novel_2024}. 

The final versions of the refined prompts, incorporating these constraints, are provided in Table~\ref{tab:all_loc_prompt_examples}.

\begin{center}
    \captionsetup{justification=centering, singlelinecheck=false}
\captionof{table}{Prompt Design and Examples for identifying all locations in disaster-related social media posts.}
\label{tab:all_loc_prompt_examples}
\begin{SectionBox}{Prompt Design for all locations}

\begin{PromptCard}[Basic Prompt — Zero Shot]{headblue}
\justifying\small
\textbf{Full prompt.}
Accurately identify all the place names (locations mentioned) in the tweet below and only respond with exact place names along with their occurrence frequency in the tweet. Keep the original spellings of place names as they appear in the tweet text and do not add any explanation in the response. Any disaster event reference, organization names, or combined hashtags such as \#Keralafloods should not be considered as place names.\\[2pt]
\textbf{Output format:}\\
\texttt{Location mentioned: Location 1 (number of occurrences), Location 2 (number of occurrences), ...}
\end{PromptCard}

\begin{PromptCard}[Persona — Zero Shot]{headviolet}
\justifying\small
\textbf{Full prompt.}
Act as an NER that recognizes all locations worldwide. Your task is to respond with exact location names along with their occurrence frequency in the tweet. Keep the original spellings of place names as they appear in the tweet text and do not add any explanation in the response. Any disaster event reference, organization names, or combined hashtags such as \#Keralafloods should not be considered as place names.\\[2pt]
\textbf{Output format:}\\
\texttt{Location mentioned: Location 1 (number of occurrences), Location 2 (number of occurrences), ...}
\end{PromptCard}

\begin{PromptCard}[Chain-of-Thought — Zero Shot]{headteal}
\justifying\small
\textbf{Full prompt.}
Basic Prompt + Think step-by-step and then provide only the final formatted answer.\\[2pt]
\textbf{Output format:}\\
\texttt{Location mentioned: Location 1 (number of occurrences), Location 2 (number of occurrences), ...}
\end{PromptCard}

\end{SectionBox}
\end{center}
\begin{center}
\begin{SectionBox}{Examples — Location Recognition}
\small

\textbf{\#E1 — Tweet}\\
``\#keralafloods \#chengannur My parents are still stranded in Madavana, Pandanad, Chengannur. No food or water has reached theme yet. Please help.''\\
\texttt{Locations mentioned: Chengannur (2), Madavana (1), Pandanad (1)}

\medskip

\textbf{\#E2 — Tweet}\\
RT @0000000000000 When you need where you want im ready for help , \#Yunanistan \#Greece \#PrayForGreece \#PrayForAthens\\
\texttt{Locations mentioned: Yunanistan (1), Greece (1)}

\medskip

\textbf{\#E3 — Tweet}\\
RT @0000000000000000 Kashmir \#Eaethquke 4 deaths and 100 injured \#Mirpur \#Pakistan administered \#Kashmir\\
\texttt{Locations mentioned: Kashmir (2), Mirpur (1), Pakistan (1)}

\medskip

\textbf{\#E4 — Tweet}\\
``Thinking of Kaikoura, those trapped on State highway 1, Nelson \& Wellington affected by the 7.5 quakes in North Canterbury @0000000 \#eqnz''\\
\texttt{Locations mentioned: Kaikoura (1), highway 1 (1), Nelson (1), Wellington (1), Canterbury (1)}

\medskip

\textbf{\#E5 — Tweet}\\
``The African Union (AU) has released US\$350{,}000 of emergency funding to Mozambique, Zimbabwe and Malawi in the aftermath of cyclone Idai. Mozambique, as the worst affected country, will receive US\$150{,}000 of that money.''\\
\texttt{Locations mentioned: Mozambique (2), Malawi (1), Zimbabwe (1)}

\medskip

\textbf{\#E6 — Tweet}\\
``655-unit oilsands work camp near Fort McMurray destroyed by wildfire''\\
\texttt{Locations mentioned: Fort McMurray (1)}

\end{SectionBox}
\end{center}

The prompts for extracting the impact and the impacted location were also developed through an iterative refinement process. Table~\ref{tab:Impact_prompt_examples} shows the refined impacts and impacted locations prompts.

\begin{center}
    \captionsetup{justification=centering, singlelinecheck=false}
\captionof{table}{Prompt Design and Examples for identifying disaster-related impacts and impacted locations in social media posts.}
\label{tab:Impact_prompt_examples}
\end{center}

\begin{center}
\begin{SectionBox}{Prompt Design for Impact and Impacted Location}

\begin{PromptCard}[Basic Prompt — Zero Shot]{headblue}
\justifying\small
\textbf{Full prompt.}
Analyse the given tweet to identify the types of disaster-related impact words (verbs or adjectives) and their corresponding impacted locations that require action from responders. Do not extract a location that has no direct impact associated with it or is vague (e.g., the region, suburban city). List the types of impacts and impacted locations as they are mentioned in the tweet text without adding details, descriptions, explanations, or hashtags.

\medskip
\textbf{Output format:}\\
\texttt{Types of Impact: <comma-separated list>}\\
\texttt{Impacted Location: <comma-separated list>}
\end{PromptCard}

\begin{PromptCard}[Persona — Zero Shot]{headviolet}
\justifying\small
\textbf{Full prompt.}
Act as an Emergency responder. Your task is to extract disaster-related impact words (verbs or adjectives) and their corresponding impacted locations from tweets that require action from responders. Do not extract a location that has no direct impact associated with it or is not explicit (e.g., ``the region''). List the types of impacts and impacted locations as they are mentioned in the tweet text without adding details, descriptions, explanations or hashtags.

\medskip
\textbf{Output format:}\\
\texttt{Types of Impact: <comma-separated list>}\\
\texttt{Impacted Location: <comma-separated list>}
\end{PromptCard}

\begin{PromptCard}[Chain-of-Thought — Zero Shot]{headteal}
\justifying\small
\textbf{Full prompt.}
Think step by step:+ Basic Prompt

\medskip
\textbf{Output format:}\\
\texttt{Types of Impact: <comma-separated list>}\\
\texttt{Impacted Location: <comma-separated list>}
\end{PromptCard}

\end{SectionBox}
\end{center}

\begin{center}
\begin{SectionBox}{Examples — Impact and Impacted Location}
\small

\textbf{\#E1 — Tweet}\\
``More than 2500 people died in Uttarakhand flood no one bats an eye! Over 250 people lose their lives in Kerala floods and everyone loses their minds! Fun Fact: North East India is currently also flooded!''\\
\texttt{Types of Impact:} died, lose their lives, flooded\\
\texttt{Impacted Location:} Uttarakhand, Kerala, North East India

\medskip

\textbf{\#E2 — Tweet}\\
``\#Greece PM Tsipras: I accept political responsibility for the tragedy in Mati. He calls on ministers to do the same. - And? That's all? No resignation for at least 87 dead people?''\\
\texttt{Types of Impact:} dead\\
\texttt{Impacted Location:} Mati

\medskip

\textbf{\#E3 — Tweet}\\
``At least 300 more people are feared dead in \#Chimanimani \& \#Chipinge due to the devastating effects of \#CycloneIdai that swept through the country over the weekend, leaving thousands homeless \& property damaged.''\\
\texttt{Types of Impact:} feared dead, homeless, property damaged\\
\texttt{Impacted Location:} Chimanimani, Chipinge

\medskip

\textbf{\#E4 — Tweet}\\
``RT @0000000000000 \#ItalyEarthquake Death toll in Amatrice alone now at 86, reports Bill Neely, NBC News''\\
\texttt{Types of Impact:} Death\\
\texttt{Impacted Location:} Amatrice

\medskip

\textbf{\#E5 — Tweet}\\
``St.\ Johns River at Main Street Bridge is in minor flood stage right now as high tide approaches.''\\
\texttt{Types of Impact:} flood\\
\texttt{Impacted Location:} St.\ Johns River at Main Street Bridge

\medskip

\textbf{\#E6 — Tweet}\\
``For those who doubted @00000000000000 video of the bus stuck in \#Chipinge I saw it, it stuck in mushy tar, the road is broken so it gonna be stuck for a while \textgreater\textgreater\textgreater\ Zimbabweans struggle with storm floods \textbar{} Al Jazeera \#CycloneIdai''\\
\texttt{Types of Impact:} CycloneIdai, stuck, broken\\
\texttt{Impacted Location:} Chipinge

\end{SectionBox}
\end{center}

For impacts, we constrained outputs to core verbs or adjectives (e.g., \textit{dead}), avoiding descriptive (e.g.,\textit{ people dead}) or numerical phrases (e.g., \textit{100 dead}). For impacted locations, we excluded vague or contextual terms (e.g., \textit{surrounds}) and restricted extraction to explicit, named locations. 

We chose “Act as an emergency responder” in the persona prompt so that the model is immediately guided to think like someone whose primary concern is identifying actionable disaster-related information. Emergency responders don’t want general descriptions or background context. They need direct, operationally relevant details, and this framing keeps the extraction task tightly focused.

\subsection{Fine-tuned Large Language Models(LLMs)}

\begin{figure}[htbp]
    \centering
    \captionsetup{justification=centering, singlelinecheck=false}
    \includegraphics[width=1\textwidth]{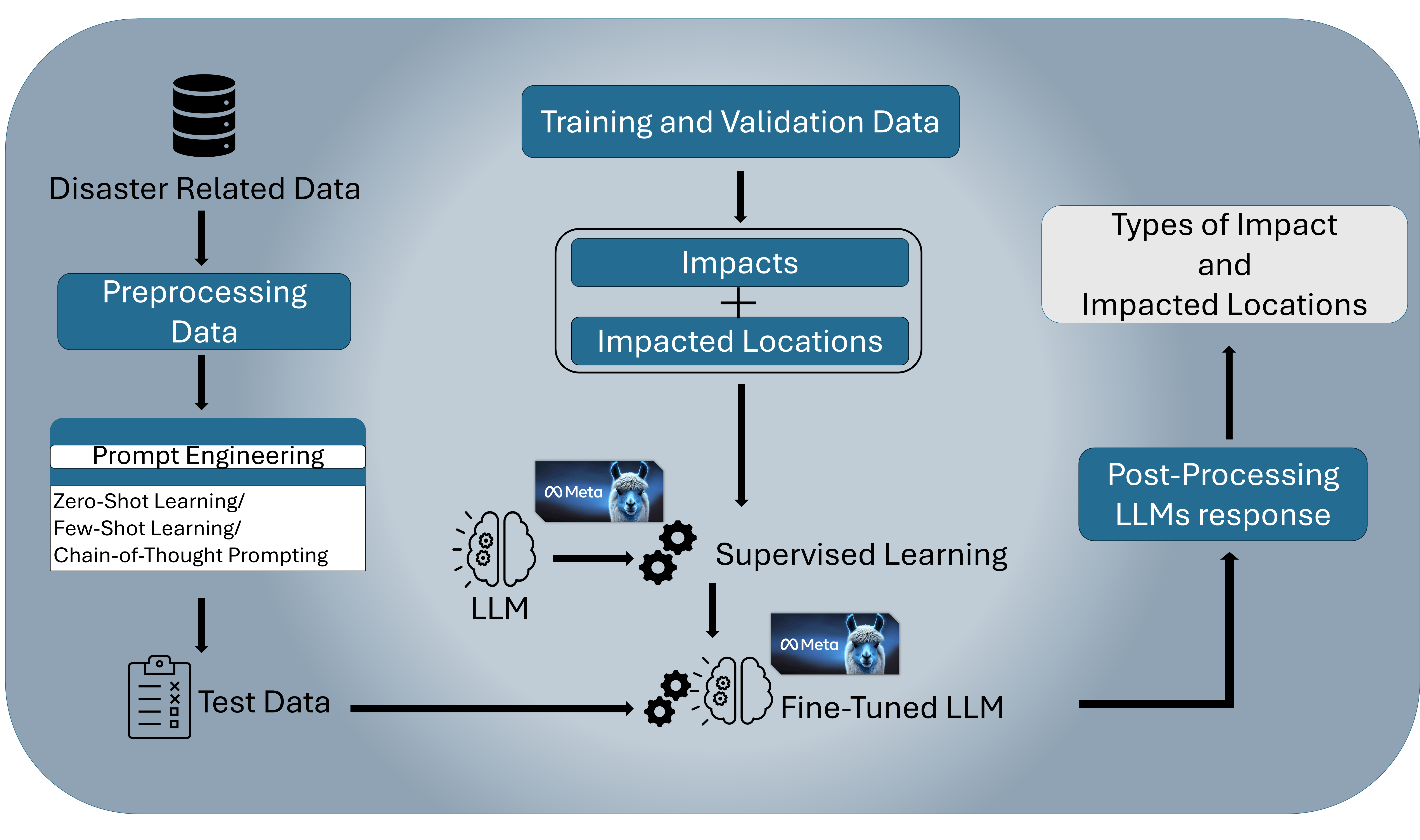}
    \caption{Workflow for fine-tuning of LLMs}
    \label{fig:LLM-flow}
\end{figure}

In addition to using the pre-trained version of the LLMs, we fine-tuned the best-performing model (Llama) identified in our evaluations in Section \ref{sec6-results}. Llama 3.3 70b was infeasible due to GPU resource constraints. To overcome this limitation, we chose to fine-tune a more accessible model, Llama 3.2 3b, specifically using the persona prompt variant. Figure \ref{fig:LLM-flow} shows the approach followed for the LLMs finetuning.

 This also reflects a practical design goal of our research, to explore methods that can operate effectively in lower-resourced environments and deliver results swiftly, which is relevant for real-time disaster impact and impacted location mapping, where computational efficiency and timely deployment are important. The fine-tuning process employed LoRA (Low-Rank Adaptation) to efficiently adapt the most critical parameters of the LLMs while freezing the majority of model weights. This method inserts trainable low-rank matrices into specific transformer layers, allowing the model to focus on learning task-specific knowledge without the computational cost of full fine-tuning for our task. 

\subsection{Proposed Approach}
\label{sec:proposed-approach}

LLMs sometimes generate content not present in the input data, which can lead to the inclusion of extraneous locations not found in the original text, known as hallucination \cite{braverman2020calibration, wang2023aligning}. LLMs do not follow a token-level labelling approach, unlike traditional NER tasks \cite{lample2016neural, wang_gpt-ner_2023}. Existing studies have sought to reduce hallucinations through retrieval-augmented generation (RAG), rubric-based self-evaluation, and internal verification techniques such as chain-of-verification or Yes/No self-checking \cite{huang2025survey, min2023factscore, chern2023evaluating, ji2023survey, dhuliawala2023chain}.

\begin{figure}[htbp]
    \centering
    \captionsetup{justification=centering, singlelinecheck=false}
    \includegraphics[width=0.5
    \textwidth]{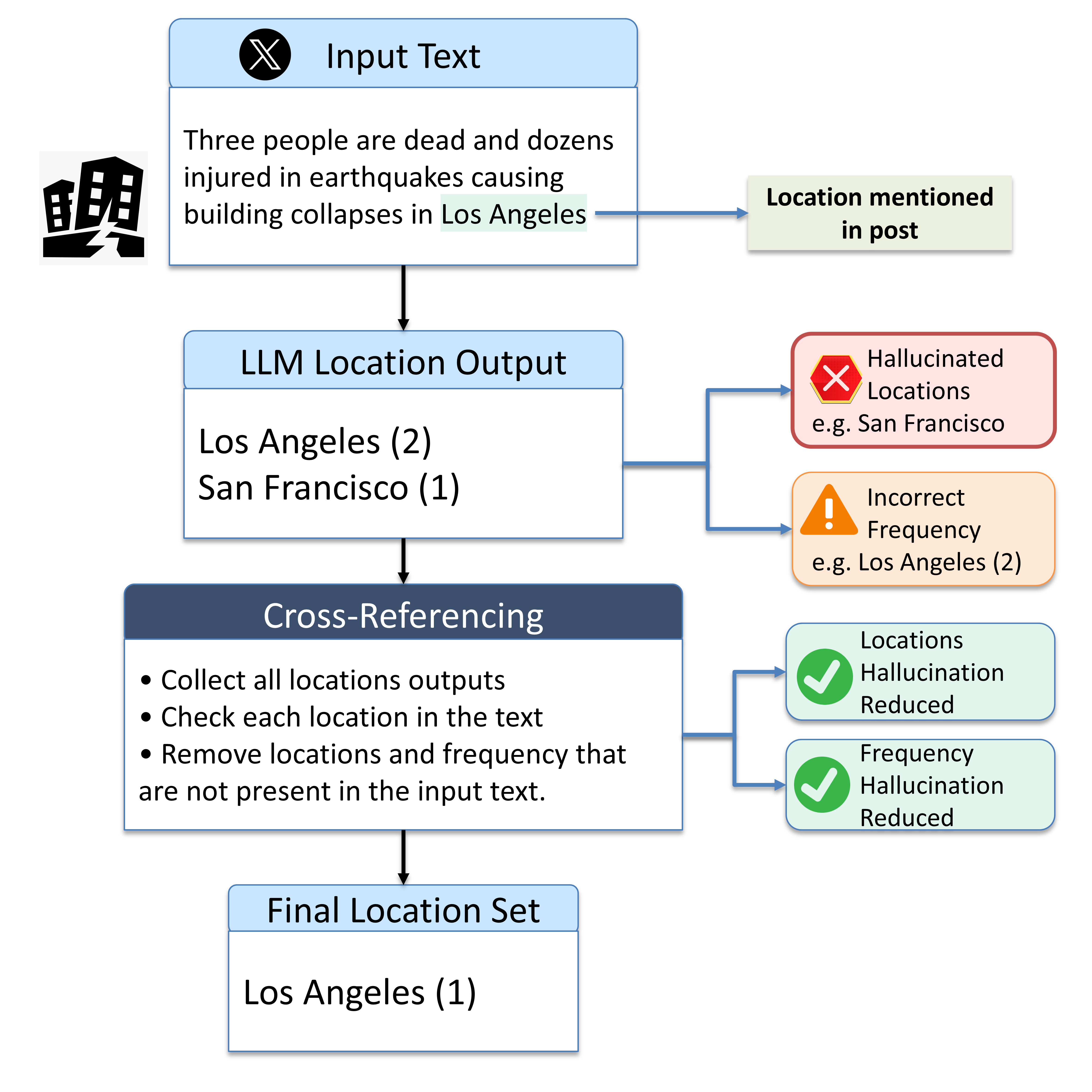} 
    \caption{Post-Processing of LLMs response}
    \label{fig:LLM-revalidation}
\end{figure}

In our experiments, LLMs sometimes misrepresented the frequency of locations present in the original text, outputting a single instance even when multiple occurrences existed and vice versa. To address these challenges, we developed a novel post-processing approach designed to mitigate hallucinations and ensure accurate frequency representation, illustrated in Fig. \ref{fig:LLM-revalidation} for the all locations task. 

The post-processing removes hallucinated outputs by validating all predicted entries against the original text. Unlike LLM-based self-verification, this approach directly compares model outputs with the source text to ensure factual accuracy. In the all location recognition task, post-processing involves checking both the presence and frequency of each predicted location. If a location or its frequency is fabricated by the LLM, either the location is removed or the frequency reduced to match that in the tweet, in the model’s output. For the impacted-location task, post-processing also verifies that all predicted impacts and impacted locations appear in the original text; however, it does not perform frequency checking.
This approach is applied to reduce hallucinations and improve the reliability of LLM outputs, rather than to modify the model’s actual predictions.

\section{Experimental Setup} \label{Sec-5:ExperimentalSetup}

This section outlines the experimental framework. We designed three experiments to address different aspects of the task, ranging from identifying all locations in disaster-related posts to distinguishing between impacted and non-impacted locations and the impacts. The details of all \textbf{Experiments 1, 2, and 3}, including models used and specific objectives, are summarised in Table~\ref{tab:experiment_overview}.

We identified the best-performing parameter settings through initial experimentation. The settings for the pre-trained LLM models were configured with temperature = 0 to ensure reproducible outputs, essential for factual extraction and top p = 0.9 to capture subtle linguistic variations in all experiments. The Llama 3.2 3b model, implemented using the Unsloth framework, was used for fine-tuning. Key hyper parameters were set to the following: learning rate 2e-4, batch size 8, LoRA rank 16, and LoRA alpha 16. The maximum number of training steps was set to 80, based on early observations of the model.

The models were evaluated based on their precision, recall, and F1-score in identifying the correct locations, impacted locations or impacts (depending on the experiment, see Table \ref{tab:experiment_overview})within the social media posts. Additionally, we assessed the models' ability to reduce false positives, particularly in distinguishing non-impacted locations from those that were directly affected by the disaster. There can be one or more predicted locations that are impacted, and these may fully or partially match the actual location mentions. So, our evaluation is based on a classification-based evaluation framework where if predicted and actual locations match, it is considered a True Positive (TP), if the location predicted by the model was not the actual location, it is assigned as a False Positive (FP), and if the actual location is not predicted at all, we consider it as a False Negative (FN), consistent with prior disaster-domain evaluations \cite{suwaileh_idrisi-d_2023a}. In cases where multiple locations are present, each predicted location is evaluated separately. If one location is correctly predicted, it contributes to TP, while other incorrectly predicted or missing locations contribute to FP or FN, respectively.

\begin{table}[H]
\centering
\captionsetup{justification=raggedright, singlelinecheck=false}
\caption{Overview of experiments, models, and purposes}
\label{tab:experiment_overview}
\resizebox{1.0\textwidth}{!}{%
\begin{tabular}{|l|l|l|l|}
\hline
\textbf{Exp.} & \textbf{Name}                                                                               & \textbf{Models Used}                                                                                    & \textbf{Purpose}                                                                                                                                                                                                                                                                                                                                         \\ \hline
1             & All Locations Recognition                                                                & \begin{tabular}[c]{@{}l@{}}Gemma-2\\ Mistral\\ Phi-3\\ Llama-3\\ Qwen-2.5\end{tabular}                  & \begin{tabular}[c]{@{}l@{}}Identify the best performing open access LLM for \\ recognition of all locations from disaster-related tweets.\end{tabular}                                                                                                                                                                                                       \\ \hline
2             & NER and LLM Comparison                                                                      & \begin{tabular}[c]{@{}l@{}}SpaCy\\ Flair\\ GLiNER\\ XLM-RoBERTa\\ BERT-NER\\ Llama-3.3 70b\end{tabular} & \begin{tabular}[c]{@{}l@{}}Compare pre-trained language models and traditional\\  NER tools with the best-performing LLM to establish \\ baseline performance for the all locations recognition task.\end{tabular}                                                                                                                                    \\ \hline
3             & \begin{tabular}[c]{@{}l@{}}Impacted Location and Impact\\  Type extraction\end{tabular} & \begin{tabular}[c]{@{}l@{}}Llama-3.3 70b-\\  (pre-trained)\\ Llama-3.2 3b \\ (fine-tuned)\end{tabular}    & \begin{tabular}[c]{@{}l@{}}Evaluate the LLM pre-trained model for identifying \\impacted  locations and impacts from disaster-related \\tweets, and  assess improvements after fine-tuning on the \\ complete disaster dataset. Furthermore, both the \\all-disaster and disaster-specific fine-tuned models were \\tested on unseen disaster types, which were not included\\ in fine-tuning, to evaluate domain specialisation.\end{tabular} \\ \hline
\end{tabular}
}
\end{table}

\section{Results and Analysis} \label{sec6-results}

The goal of \textbf{Experiment 1} was to identify the best performing open access model for the all locations task, to help us select which model to use for subsequent experiments.  The results are presented in Table \ref{tab:comparison_of_different_LLMs}, which indicates that Llama 3.3-70b outperforms other state-of-the-art LLMs with an F1-score of 0.77, while Qwen2.5-7b, Gemma2-9b, Mistral-7b and phi3-14b achieved F1-scores ranging from 0.60 to 0.73. Thus, Llama 3.3 70b was selected for Experiment 2. 

In \textbf{Experiment 2}, we compared traditional NER tools and pre-trained NER models with the best-performing LLM to determine which approach performs best at identifying every location mentioned in disaster-related social media posts, regardless of whether the location was impacted or not.
Table \ref{tab:Comparison_of_all_Location_Recognition_models} indicates SpaCy performed the best among all pre-trained NER models, with an F1-score of 0.81. Both Flair and GLiNER followed, each exhibiting a performance of F1-score 0.79. However, the Llama 3.3 70b model outperforms all NER tools, with F1-scores between 0.84 and 0.86 depending on prompt. The superior performance of Llama 3.3 70b over pre-trained NER models confirms the dominance of LLMs for NER through their vast contextual knowledge and instruction-following capability. These results align with recent studies that show LLMs outperform traditional NER models in different domains \cite{hu_geo-knowledge-guided_2023, obeidat2025llms,xiao_llm-named_2024}. 
Table \ref{tab:Comparison_of_all_Location_Recognition_models} results also demonstrate that the proposed post-processing approach significantly enhances the effectiveness of the LLMs in Experiments 2.

\begin{table}[H]
\centering
    \captionsetup{justification=centering, singlelinecheck=false}
\caption{Experiment 1: Initial comparison of different LLMs on a small dataset without Post-Processing (-PP)}
\label{tab:comparison_of_different_LLMs}
\resizebox{0.7\columnwidth}{!}{%
\begin{tabular}{|l|l|l|l|l|l|}
\hline
\textbf{Model} & \textbf{Version} & \textbf{Prompt\_Type} & \textbf{Precision} & \textbf{Recall} & \textbf{F1-score} \\ \hline
\textbf{} & \textbf{} & \textbf{} & \textbf{-PP} & \textbf{-PP} & \textbf{-PP} \\ \hline
\textbf{Qwen2.5} & 7b & Basic zero-shot & 0.66 & 0.83 & 0.73 \\ \hline
\textbf{Phi3} & 14b & Basic zero-shot & 0.65 & 0.71 & 0.68 \\ \hline
\textbf{Gemma2} & 9b & Basic zero-shot & 0.59 & 0.86 & 0.70 \\ \hline
\textbf{Mistral} & 7b & Basic zero-shot & 0.49 & 0.79 & 0.60 \\ \hline
\textbf{Llama3.3} & 70b & Basic zero-shot & 0.66 & 0.92 & \textbf{0.77} \\ \hline
\end{tabular}
}
\end{table}

\begin{table}[H]
\centering
    \captionsetup{justification=centering, singlelinecheck=false}
\caption{Experiment 2: Performance Comparison of all Location Recognition Models with Post-Processing (PP) and without the Post-Processing (-PP)}
\label{tab:Comparison_of_all_Location_Recognition_models}
\resizebox{\columnwidth}{!}{%
\begin{tabular}{|l|l|l|ll|ll|ll|}
\hline
\textbf{Model} &
  \textbf{Version} &
  \textbf{} &
  \multicolumn{2}{l|}{\textbf{Precision}} &
  \multicolumn{2}{l|}{\textbf{Recall}} &
  \multicolumn{2}{l|}{\textbf{F1-score}} \\ \hline
\textbf{GLiNER} &
  gliner 2.13 &
   &
  \multicolumn{1}{l|}{} &
  0.78 &
  \multicolumn{1}{l|}{} &
  0.80 &
  \multicolumn{1}{l|}{} &
  0.79 \\ \hline
\textbf{Spacy} &
  en\_core\_web\_trf &
   &
  \multicolumn{1}{l|}{} &
  0.85 &
  \multicolumn{1}{l|}{} &
  0.78 &
  \multicolumn{1}{l|}{} &
  0.81 \\ \hline
\textbf{Flair} &
  flair 15.0 &
   &
  \multicolumn{1}{l|}{} &
  0.78 &
  \multicolumn{1}{l|}{} &
  0.80 &
  \multicolumn{1}{l|}{} &
  0.79 \\ \hline
\textbf{XLM-RoBERTa} &
  large-finetuned-conll03-english &
   &
  \multicolumn{1}{l|}{} &
  0.62 &
  \multicolumn{1}{l|}{} &
  0.79 &
  \multicolumn{1}{l|}{} &
  0.70 \\ \hline
\textbf{BERT NER} &
  dslim/bert-base-NER &
   &
  \multicolumn{1}{l|}{} &
  0.67 &
  \multicolumn{1}{l|}{} &
  0.77 &
  \multicolumn{1}{l|}{} &
  0.72 \\ \hline
\textbf{} &
  \textbf{} &
  \textbf{Prompt\_Type} &
  \multicolumn{1}{l|}{\textbf{-PP}} &
  \textbf{PP} &
  \multicolumn{1}{l|}{\textbf{-PP}} &
  \textbf{PP} &
  \multicolumn{1}{l|}{\textbf{-PP}} &
  \textbf{PP} \\ \hline
\textbf{Llama3} &
  3.3, 70B &
  Basic zero-shot &
  \multicolumn{1}{l|}{0.66} &
  0.78 &
  \multicolumn{1}{l|}{0.92} &
  0.92 &
  \multicolumn{1}{l|}{0.77} &
  0.85 \\ \hline
\textbf{Llama3} &
  3.3, 70B &
  Basic one-shot &
  \multicolumn{1}{l|}{0.68} &
  0.80 &
  \multicolumn{1}{l|}{0.93} &
  0.92 &
  \multicolumn{1}{l|}{0.78} &
  \textbf{0.86} \\ \hline
\textbf{Llama3} &
  3.3, 70B &
  Basic 6-shot &
  \multicolumn{1}{l|}{0.69} &
  0.81 &
  \multicolumn{1}{l|}{0.91} &
  0.90 &
  \multicolumn{1}{l|}{0.78} &
  0.85 \\ \hline
\textbf{Llama3} &
  3.3, 70B &
  Persona zero-shot &
  \multicolumn{1}{l|}{0.67} &
  0.79 &
  \multicolumn{1}{l|}{0.91} &
  0.91 &
  \multicolumn{1}{l|}{0.77} &
  0.84 \\ \hline
\textbf{Llama3} &
  3.3, 70B &
  Persona one-shot &
  \multicolumn{1}{l|}{0.67} &
  0.80 &
  \multicolumn{1}{l|}{0.92} &
  0.92 &
  \multicolumn{1}{l|}{0.78} &
  0.85 \\ \hline
\textbf{Llama3} &
  3.3, 70B &
  Persona 6-shot &
  \multicolumn{1}{l|}{0.68} &
  0.80 &
  \multicolumn{1}{l|}{0.92} &
  0.91 &
  \multicolumn{1}{l|}{0.78} &
  0.85 \\ \hline
\textbf{Llama3} &
  3.3, 70B &
  COT zero-shot &
  \multicolumn{1}{l|}{0.67} &
  0.79 &
  \multicolumn{1}{l|}{0.92} &
  0.91 &
  \multicolumn{1}{l|}{0.78} &
  0.85 \\ \hline
\textbf{Llama3} &
  3.3, 70B &
  COT one-shot &
  \multicolumn{1}{l|}{0.62} &
  0.76 &
  \multicolumn{1}{l|}{0.92} &
  0.92 &
  \multicolumn{1}{l|}{0.74} &
  0.84 \\ \hline
\end{tabular}%
}
\end{table}

We evaluated the Llama 3.3 70b pre-trained model using various prompt engineering strategies of zero-shot to 6-shot with Basic, Persona, and Chain-of-Thought (COT) prompts. The Basic zero-shot Prompt achieved an F1-score of 0.85, and Basic One-Shot Prompt achieved the highest F1-score of 0.86 in identifying all locations. This suggests that when adding more examples, or even using alternate example configurations (which were also tested), the performance remains lower in the 6-Shot prompt compared to One-Shot. This observation is consistent with the findings of \citet{zhao2021calibrate}, who noted that few-shot prompts often introduce label bias due to imbalanced examples, whereas carefully calibrated one-shot prompts can avoid this bias and yield more accurate and reliable outputs.
XLM-RoBERTa, which is a multi-lingual model based on the RoBERTa architecture, demonstrated lower performance than Flair and GLiNER with a F1-score of 0.70. Although \citet{kopanov_comparative_2024} found XLM-RoBERTa performed well for general NER tasks, its reduced performance in the disaster specific location recognition task suggests that it may lack the specialised contextual understanding required for such domain-specific tasks, due to its pre-training on general-purpose datasets.
BERT-NER, although trained on location data, achieved a low F1-score of 0.72 due to its limited contextual understanding to reason beyond token-level boundaries, and difficulty in handling the informal and noisy language often present in disaster-related tweets.

The most successful LLM was fine-tuned and applied with persona-based training with 6-shots and post-processing to identify impacted locations and types of impact within the nuanced language of disaster-related social media posts in \textbf{Experiment 3}. Table \ref{tab: Comparison of pre-trained and fine-tuned Llama models} shows the results of impacts and impacted locations extraction task. On the DILC, the pre-trained Llama 3.2 3b model achieved an F1 score of 0.49 for impacts and 0.73 for impacted locations, whereas the variant that was fine-tuned on 68\% randomly selected training data and 20\% randomly selected testing data without overlapping improved 0.69 for impacts while maintaining a stable 0.74 for impacted locations. 

We further evaluated disaster-specific subsets of tweets written in English from two countries (2017 Hurricane Harvey\footnote{\url{https://en.wikipedia.org/wiki/Hurricane_Harvey}} from the United States and the 2016 Kaikoura earthquake from New Zealand\footnote{\url{https://en.wikipedia.org/wiki/2016_Kaikōura_earthquake}}) where English is an official language; and two countries (the 2019 Pakistan earthquake\footnote{\url{https://en.wikipedia.org/wiki/2019_Kashmir_earthquake}} and the 2018 Greece wildfires\footnote{\url{https://en.wikipedia.org/wiki/2018_Attica_wildfires}}) where it is not, and the benefits of fine-tuning became more pronounced. These disaster specific subsets were derived from the DILC dataset and were selected based on the diversity of disaster types, the volume of available tweets for training and testing, and the variation between native and non native English speaking countries. For each target disaster, a corresponding domain-specific dataset was created, i.e, DILC-E (which contains all tweets that reference an earthquake in the DILC), DILC-W (wildfires tweets from DILC), and DILC-H (hurricane tweets from DILC). From each of these data sets, the tweets that refer to the particular disaster that is being tested were excluded.  That is, tweets referring to Hurricane Harvey were removed from DILC-H, and similarly, tweets referring to the 2018 Greece Wildfires were removed from DILC-W. Two variations of DILC-E were formulated, each excluding tweets for the respective disaster.  That is, DILC-EN excludes tweets from the 2016 Kaikoura Earthquake and DILC-EP excludes tweets from the 2019 Pakistan Earthquake. Details of fine-tuning and test datasets and their sizes are provided in Table \ref{tab: Comparison of pre-trained and fine-tuned Llama models}

Two fine-tuning settings were applied. First, an all-disaster model was fine-tuned on tweets from multiple disaster types, excluding the target disaster (e.g., DILC without 2016 Kaikoura Earthquake tweets (NZ)) and then tested on the excluded disaster’s tweets. Second, a disaster-specific model was fine-tuned on tweets of the same disaster type (e.g., DILC-EN excluding 2016 Kaikoura Earthquake tweets) and then tested on the target event’s tweets. Both models were evaluated on unseen disaster data excluded from their respective fine-tuning sets to assess domain specialisation and generalisation across disaster contexts.

For impact extraction, the best performance was observed on the Pakistan Earthquake dataset, where fine-tuning on earthquake data yielded an F1 score of 0.86, compared to 0.83 when fine-tuned on the entire DILC data set (excluding the Pakistan Earthquake tweets). This highlights the effectiveness of disaster-specific fine-tuning in capturing disaster-related linguistic cues for the detection of impacts. This reflects the tendency for some impacts to be specific to certain disaster types (for example, an earthquake might cause a bridge to collapse, but is less likely to cause fire-related impacts).

For impacted location extraction, the highest performance was obtained on the Greece Wildfire dataset, where the model fine-tuned on all disaster data achieved an F1 score of 0.77. This suggests that exposure to a diverse range of disaster-related texts enhances the model’s ability to generalise across unseen disaster scenarios when detecting impacted locations.

On all disasters (DILC), the fine-tuned 3b model also surpasses both the pre-trained 3b (0.49; +0.20) and the larger pre-trained 70b (0.60; +0.09). These results further support the evidence that disaster-specific fine-tuning can improve impact extraction. In contrast, disaster-specific fine-tuning does not generally improve impacted location extraction, with negligible performance of NZ (+0.01) and Harvey (-0.01), and notable performance drops particularly for the case of Pakistan (-0.06) and Greece (-0.12).

The decline of the impacted location in disaster-specific fine-tuning is affected by both the size of the fine-tuning corpus and the number of location mentions, influencing model performance. In the test datasets, ratios of impacted locations to all location mentions vary: New Zealand includes 70/98 (71.4 \%); Harvey 163/215 (75.8 \%); Pakistan 121/244 (49.6 \%) and Greece 343/508 (67.5 \%), thereby increasing ambiguity in identifying impacted locations that describe the impacts. Another challenge arises from hierarchical location mentions, such as city, province, and country co-occur within the same tweet, making it difficult for the model to distinguish the most specific impacted location. We also observe that the model sometimes extracts impacted locations with added descriptive context (e.g., \textit{Athens} vs \textit{regions surrounding Athens}) or only partially captures locations (e.g.,\textit{Marlborough Sounds} vs \textit{Marlborough}). Additionally, the model sometimes produced hallucinatory responses where the impact is mistakenly identified as the impacted location.

\begin{table}[!h]
\centering
    \captionsetup{justification=centering, singlelinecheck=false}
\caption{Experiment 3: Performance comparison of pre-trained and fine-tuned Llama models for impacted-location and impact-type extraction (Persona 6-shots)}
\label{tab: Comparison of pre-trained and fine-tuned Llama models}
\resizebox{\columnwidth}{!}{%
\begin{tabular}{|cccc|cc|ccc|ccc|}
	\hline
	\multicolumn{4}{|c|}{\textbf{}} &
	\multicolumn{2}{c|}{\textbf{Dataset Size (Tweets)}} &
	\multicolumn{3}{c|}{\textbf{Impact}} &
	\multicolumn{3}{c|}{\textbf{Location}} \\ \hline
	\multicolumn{1}{|c|}{\textbf{Disaster}} &
	\multicolumn{1}{c|}{\textbf{Model}} &
	\multicolumn{1}{c|}{\textbf{Finetuning}} &
	\textbf{Dataset} &
	\multicolumn{1}{c|}{\textbf{Fine-tune}} &
	\textbf{Test} &
	\multicolumn{1}{c|}{\textbf{P}} &
	\multicolumn{1}{c|}{\textbf{R}} &
	\textbf{F1} &
	\multicolumn{1}{c|}{\textbf{P}} &
	\multicolumn{1}{c|}{\textbf{R}} &
	\textbf{F1} \\ \hline
	\multicolumn{1}{|c|}{\multirow{3}{*}{\textbf{All Disasters}}} &
	\multicolumn{1}{c|}{Llama 3.3} &
	\multicolumn{1}{c|}{No} &
	DILC &
	\multicolumn{1}{c|}{-} &
	300 &
	\multicolumn{1}{c|}{0.66} &
	\multicolumn{1}{c|}{0.54} &
	0.60 &
	\multicolumn{1}{c|}{0.62} &
	\multicolumn{1}{c|}{0.89} &
	0.73 \\ \cline{2-12} 
	\multicolumn{1}{|c|}{} &
	\multicolumn{1}{c|}{Llama 3.2} &
	\multicolumn{1}{c|}{No} &
	DILC &
	\multicolumn{1}{c|}{-} &
	300 &
	\multicolumn{1}{c|}{0.56} &
	\multicolumn{1}{c|}{0.44} &
	0.49 &
	\multicolumn{1}{c|}{0.65} &
	\multicolumn{1}{c|}{0.83} &
	0.73 \\ \cline{2-12} 
	\multicolumn{1}{|c|}{} &
	\multicolumn{1}{c|}{Llama 3.2} &
	\multicolumn{1}{c|}{Yes} &
	DILC &
	\multicolumn{1}{c|}{997} &
	300 &
	\multicolumn{1}{c|}{0.71} &
	\multicolumn{1}{c|}{0.66} &
	\textbf{0.69} &
	\multicolumn{1}{c|}{0.77} &
	\multicolumn{1}{c|}{0.70} &
	\textbf{0.74} \\ \hline
	\multicolumn{1}{|c|}{\multirow{2}{*}{\textbf{\begin{tabular}[c]{@{}c@{}}Kaikoura Earthquake\\ (NZ)\end{tabular}}}} &
	\multicolumn{1}{c|}{Llama 3.2} &
	\multicolumn{1}{c|}{Yes} &
	\begin{tabular}[c]{@{}c@{}}DILC \\ ex Kaikoura Earthquake\end{tabular} &
	\multicolumn{1}{c|}{1401} &
	60 &
	\multicolumn{1}{c|}{0.67} &
	\multicolumn{1}{c|}{0.72} &
	0.69 &
	\multicolumn{1}{c|}{0.67} &
	\multicolumn{1}{c|}{0.66} &
	0.66 \\ \cline{2-12} 
	\multicolumn{1}{|c|}{} &
	\multicolumn{1}{c|}{Llama 3.2} &
	\multicolumn{1}{c|}{Yes} &
	DILC-EN &
	\multicolumn{1}{c|}{252} &
	60 &
	\multicolumn{1}{c|}{0.68} &
	\multicolumn{1}{c|}{0.78} &
	0.73 &
	\multicolumn{1}{c|}{0.68} &
	\multicolumn{1}{c|}{0.66} &
	0.67 \\ \hline
	\multicolumn{1}{|c|}{\multirow{2}{*}{\textbf{\begin{tabular}[c]{@{}c@{}}Hurricane Harvey \\ (USA)\end{tabular}}}} &
	\multicolumn{1}{c|}{Llama 3.2} &
	\multicolumn{1}{c|}{Yes} &
	\begin{tabular}[c]{@{}c@{}}DILC \\ ex Hurricane Harvey\end{tabular} &
	\multicolumn{1}{c|}{1316} &
	145 &
	\multicolumn{1}{c|}{0.65} &
	\multicolumn{1}{c|}{0.66} &
	0.66 &
	\multicolumn{1}{c|}{0.74} &
	\multicolumn{1}{c|}{0.77} &
	0.76 \\ \cline{2-12} 
	\multicolumn{1}{|c|}{} &
	\multicolumn{1}{c|}{Llama 3.2} &
	\multicolumn{1}{c|}{Yes} &
	DILC-H &
	\multicolumn{1}{c|}{255} &
	145 &
	\multicolumn{1}{c|}{0.70} &
	\multicolumn{1}{c|}{0.66} &
	0.68 &
	\multicolumn{1}{c|}{0.75} &
	\multicolumn{1}{c|}{0.75} &
	0.75 \\ \hline
	\multicolumn{1}{|c|}{\multirow{2}{*}{\textbf{\begin{tabular}[c]{@{}c@{}}Pakistan Earthquake \\ (PK)\end{tabular}}}} &
	\multicolumn{1}{c|}{Llama 3.2} &
	\multicolumn{1}{c|}{Yes} &
	\begin{tabular}[c]{@{}c@{}}DILC \\ ex Pakistan Earthquake\end{tabular} &
	\multicolumn{1}{c|}{1372} &
	89 &
	\multicolumn{1}{c|}{0.92} &
	\multicolumn{1}{c|}{0.76} &
	0.83 &
	\multicolumn{1}{c|}{0.62} &
	\multicolumn{1}{c|}{0.50} &
	0.55 \\ \cline{2-12} 
	\multicolumn{1}{|c|}{} &
	\multicolumn{1}{c|}{Llama 3.2} &
	\multicolumn{1}{c|}{Yes} &
	DILC-EP &
	\multicolumn{1}{c|}{223} &
	89 &
	\multicolumn{1}{c|}{0.93} &
	\multicolumn{1}{c|}{0.79} &
	\textbf{0.86} &
	\multicolumn{1}{c|}{0.54} &
	\multicolumn{1}{c|}{0.46} &
	0.49 \\ \hline
	\multicolumn{1}{|c|}{\multirow{2}{*}{\textbf{\begin{tabular}[c]{@{}c@{}}Greece WildFire\\ (GR)\end{tabular}}}} &
	\multicolumn{1}{c|}{Llama 3.2} &
	\multicolumn{1}{c|}{Yes} &
	\begin{tabular}[c]{@{}c@{}}DILC \\ ex Greece Wildfire\end{tabular} &
	\multicolumn{1}{c|}{1149} &
	312 &
	\multicolumn{1}{c|}{0.71} &
	\multicolumn{1}{c|}{0.65} &
	0.68 &
	\multicolumn{1}{c|}{0.78} &
	\multicolumn{1}{c|}{0.77} &
	\textbf{0.77} \\ \cline{2-12} 
	\multicolumn{1}{|c|}{} &
	\multicolumn{1}{c|}{Llama 3.2} &
	\multicolumn{1}{c|}{Yes} &
	DILC-W &
	\multicolumn{1}{c|}{70} &
	312 &
	\multicolumn{1}{c|}{0.79} &
	\multicolumn{1}{c|}{0.73} &
	0.76 &
	\multicolumn{1}{c|}{0.66} &
	\multicolumn{1}{c|}{0.63} &
	0.65 \\ \hline
\end{tabular}%
}
\end{table}

Overall, the LLM models outperformed traditional NER models that are pre-trained on location entities for the all location recognition task. Tailored prompting strategies and effective post-processing further improved the prediction of the all locations recognition task. The comparative evaluation of the all-disaster and disaster-specific fine-tuned models across different disaster datasets demonstrates distinct strengths. The disaster-specific fine-tuning yielded superior performance in identifying impacts, reflecting the advantages of domain-focused learning. In contrast, the all-disaster fine-tuning achieved the best results in impacted location extraction, indicating that exposure to diverse disaster contexts enhances understanding of location information. However, the comparatively lower performance in disaster-specific fine-tuning also underscores the need for further methodological advances, such as incorporating location resolution and hierarchical place representations, to achieve robust and reliable extraction of impacted locations.

\section{Discussion} \label{sec7-discussion}

The results reveal varying strengths and limitations of LLMs in handling diverse disaster types, information categories, and linguistic complexities. One recurring challenge was span variation, which often led to partial matches in both the impacts and the impacted locations. That is, models frequently predicted only the base form of an impact or impacted location, while the gold standard included an extended form (e.g., \textit{Jatlaan} vs. \textit{Jatlaan Canal}, \textit{found dead} vs. \textit{dead}), or vice versa. Future work should therefore employ span-based metrics and consider overlap-based measures such as Soft Jaccard to provide partial credit for nearly missed predictions. Additional difficulties arose from granularity mismatches, such as when multiple locations were mentioned in hierarchical order (e.g., \textit{Jhelum, Pakistan} two location levels, but only the city \textit{Jhelum} represents the directly impacted area.) or when locations of different levels mentioned separately (e.g., \textit{\#Athens \#Greece}), often leading to confusion in selecting the appropriate level of specificity. These inconsistencies reflect the high variability and complexity of location usage in disaster contexts.

\section{Conclusion and Future Work} \label{sec8-conclusion}

Social media provides real-time, ground-level information during disasters, often revealing critical details about affected locations and impacts. This study introduced LLMs to extract types of impacts and impacted locations from disaster-related posts. For this purpose we developed the DILC Corpus, based on 19 major natural disasters spanning 11 countries, covering both native and non-native English speaking regions.

The results show that LLMs, particularly Llama, outperform conventional NER approaches when guided by context-aware prompts and disaster-specific data. Despite challenges such as span variation and granularity mismatches, fine-tuning led to improvements across disaster types. The findings highlight the value of domain adaptation and structured post-processing for reliable extraction of actionable information. Overall, this work demonstrates the potential of LLMs to transform unstructured social media content into structured insights for enhanced situational awareness and decision support in disaster response.

In the future, we aim to extend the current framework by incorporating multilingual datasets, thereby improving the models’ adaptability across diverse linguistic and cultural contexts. In parallel, geocoding extracted locations for precise resource targeting and developing granular impact information on severity and affected entities (e.g., \textit{buildings, roads, people}) will further elevate the practical utility of LLMs in disaster preparedness, response, and recovery \cite{francis2022annotating}. By addressing these directions, LLMs can be further advanced into robust, end-to-end systems for comprehensive disaster information management.

\vspace{-\baselineskip}

\section*{Funding}

This project was partially supported by QuakeCoRE, a New Zealand Tertiary Education Commission-funded Centre of Research Excellence: QuakeCoRE publication number 1108 and partially by Massey University.

\bibliographystyle{unsrtnat}
\bibliography{PAPERS}

\clearpage

\appendix


\section*{Supplementary material}

Appendix A. Annotation Guidelines

## *Annotation Guidelines*

Everyone tags 300 common tweets and 612 individually assigned using the following scheme:

**Basic Tags:**

Disaster impact information:
- **Type of impact**
  - We are looking for text that identifies the type of damage or impact that the item has encountered. For example, this could include malfunction, closure, or physical damage. Impact type (e.g., "death", "injury", "damage to buildings", "misplaced") is even mentioned as a hashtag.
  - Negative statements about impacts (road not damaged) should also be tagged, in which case both words (not damaged) should be tagged. If the tweet contains "destroy and damage" together in a sentence, it should be "destroyed" and "damage" as separate impacts.
  - In case of "death" vs "death toll", keep only death as the impact.
  - Main disasters, Flash floods, wildfires, and earthquakes, etc should be highlighted as the type of impact when associated with the locations.

  **Example Annotations:**
  - "many lost their family/crops" → Annotate **"loss" as a type of impact**
  - The death toll is rising to 247" → Annotate **"death"** as impact
  - At least 17 homes on Black Creek were destroyed. 47 sustained major damage. By @00000000000" → Annotate **destroyed, damage as " impact "**

- **Location information:**
- **Impacted location**
  - Only tag items that describe the location of the impact
  - Identify text that describes the location of the item damaged and the disaster impact area. This could be a city, town or a smaller location area.

  For Example, *"Jatlaan canal embankment is damaged. It is one of the main canal originating from Mangla dam which irrigates Punjab. #earthquake AJK Mirpur"* contains multiple place names. However, only the "Jatlaan canal" is directly impacted, where 'damaged' is the impact and 'canal embankment' is the item affected.

  - Impacted Location mentions within the post, as either misspelt or as an abbreviation/short form (NZ for New Zealand) must also be tagged.
  - Impacted locations mentioned in hashtags such as #dallas #texas will also be tagged without including a hashtag sign. Impacted locations mentioned in hashtags, along with other words, will not be tagged, such as #prayfordallas
  - Buildings such as hotels should be items affected unless it is part of the location, for example, "DoubleTree Hotel Orlando Airport".

***Which Tweets to Annotate:***

The 1525 will be drawn from the following three classes in the IDRISI data sets for annotation:
**-infrastructure and utilities damage**
**-injured or dead people**
**-Missing-or-found**
Each person should do:
- 300 in common from all disasters (all people tag the same 300 tweet
- After the first 300 tweet annotations, we calculate IAA.
- The rest of the 612 tweets can be annotated afterwards.

Meta-Annotator will use Python code to randomly select 300 entries and distribute the same set among annotators. After addressing any corrections and meeting the requirements, the remaining 612 entries will be distributed for annotation. Once these entries are annotated, the Inter-Annotator Agreement (IAA) will be calculated.

***Tagging tool:***
**Brat tag annotating instructions**
Tag types for the data.

**For entity tagging**
**Tagging Entities**
You will use the Brat interface to highlight and tag parts of the text as specific entities.
**Steps for tagging an entity**:
1. **Select the Text**: Highlight the portion of the text that you want to tag as an entity.
2. **Choose the Entity Type**: Once the text is selected, a menu will appear where you can choose the entity type. The available entity types for this project are:
   - **impacttype**
   - **impactedlocation**
3. **Save the Tag**: After selecting the entity type, click to confirm. The tagged entity will appear as highlighted in the text, with the appropriate label attached.
4. **Delete/change Tag**: Double-click the entity label, a dialogue box will appear with the list of labels, select the delete option to delete.

**Save the Entities**: Once the entities are tagged, confirm the action by saving it.



\end{document}